\documentclass{article}
\usepackage{spconf,amsmath,graphicx,hyperref,booktabs,tabularx,multirow,amssymb}
\usepackage{dsfont}             
\usepackage{float}
\usepackage[caption=false]{subfig}   
\usepackage[english]{babel}     
\usepackage{microtype}          
\usepackage{soul}
\newcommand{\uline}[1]{\ul{\mbox{#1}}}  

\title{GenLie: A Global-Enhanced Lie Detection Network under Sparsity and Semantic Interference}
\name{Zongshun Zhang$^{\star}$\thanks{
        This research was supported by the National Natural Science Foundation of China (U22B2061), the National Key R\&D Program of China (2022YFB4300603).
    }, Yao Liu$^{\star}$, Qiao Liu$^{\star}$, Xuefeng Peng$^{\star}$, Peiyuan Jiang$^{\star}$, \emph{Jiaye Yang}$^{\star}$, \emph{Daibing Yao}$^{\dagger}$, \emph{Wei Lin}$^{\dagger}$}

\address{
    $^{\star}$ School of Computer Science and Engineering, \\
    University of Electronic Science and Technology of China, Sichuan, China \\
    $^{\dagger}$ Yizhou Prison, Sichuan, China
}

\begin{document}
%
\maketitle
\begin{abstract}
Video-based lie detection aims to identify deceptive behaviors from visual cues. Despite recent progress, its core challenge lies in learning sparse yet discriminative representations. Deceptive signals are typically subtle and short-lived, easily overwhelmed by redundant information, while individual and contextual variations introduce strong identity-related noise. To address this issue, we propose \textbf{GenLie}, a \textbf{G}lobal-\textbf{En}hanced \textbf{Lie} Detection Network that performs local feature modeling under global supervision. Specifically, sparse and subtle deceptive cues are captured at the local level, while global supervision and optimization ensure robust and discriminative representations by suppressing identity-related noise. Experiments on three public datasets, covering both high- and low-stakes scenarios, show that GenLie consistently outperforms state-of-the-art methods. Source code is available at \href{https://github.com/AliasDictusZ1/GenLie}{https://github.com/AliasDictusZ1/GenLie}.
\end{abstract}
\section{Introduction}
Deception frequently occurs in daily communication as well as in high-stakes domains such as courtroom interrogations, security screenings, financial investigations, and competitive game shows \cite{burzo2018multimodal,porter2010truth,gneezy2005deception,chen2023trust}. Yet, human lie detection remains unreliable, with an accuracy of only about 54\% and further susceptibility to subjective biases \cite{bond2006accuracy,depaulo2003cues}.
These limitations motivate the development of video-based lie detection systems capable of learning robust and discriminative representations.  

\begin{figure}
    \centering
    \includegraphics[width=0.96\linewidth]{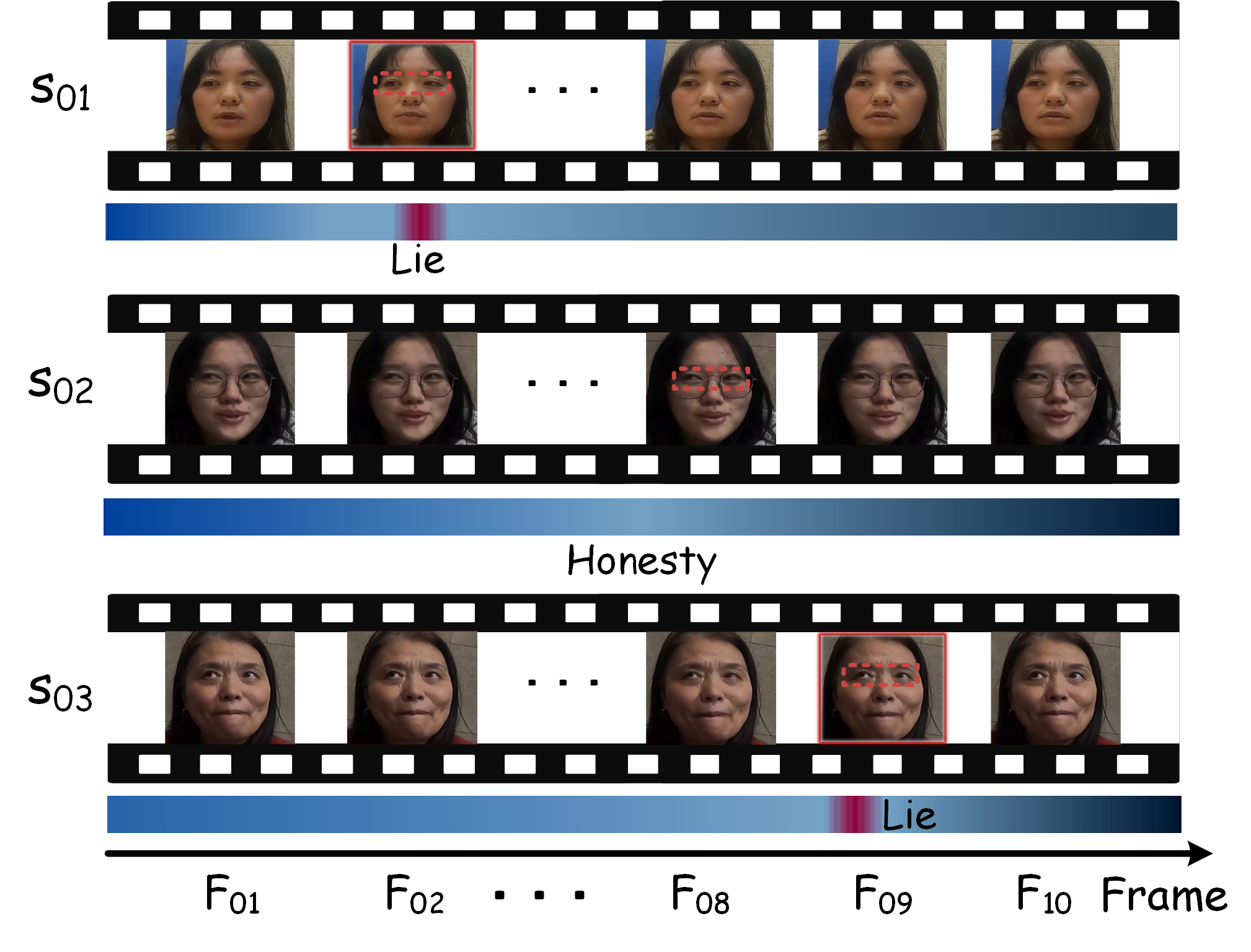}
    \caption{Core challenge in video-based lie detection: subtle and transient cues are often overwhelmed by noise and contextual bias.}
    \label{fig:motivation}
\end{figure}

Recent approaches to video-based lie detection have shifted from hand-crafted psychological cues to deep neural networks capable of end-to-end feature learning. Existing methods can be broadly categorized into explicit, implicit, and hybrid modeling strategies. Explicit approaches rely on predefined psychological cues such as micro-expressions or action units \cite{yildirim2023influence}, which are interpretable but easily miss subtle or undefined deceptive behaviors. Implicit approaches employ CNNs or Transformers to automatically learn spatiotemporal patterns from raw video \cite{lin2019tsm,wang2023rethinking}, achieving promising results but struggling with sparse signals and semantic interference. Hybrid approaches combine handcrafted and deep features for improved robustness \cite{ding2019face,nam2023facialcuenet}, but often introduce complex multi-branch architectures that limit scalability. Overall, video-based lie detection remains fundamentally constrained by the challenge of learning sparse yet discriminative representations.
This difficulty is due to deceptive behaviors are typically subtle, short-lived, and scattered across temporal segments (Fig.~\ref{fig:motivation}), making them easily overwhelmed by redundancy and entangled with identity- or context-related variations.

\begin{figure*}[h]
  \centering
  \includegraphics[width=0.75\textwidth]{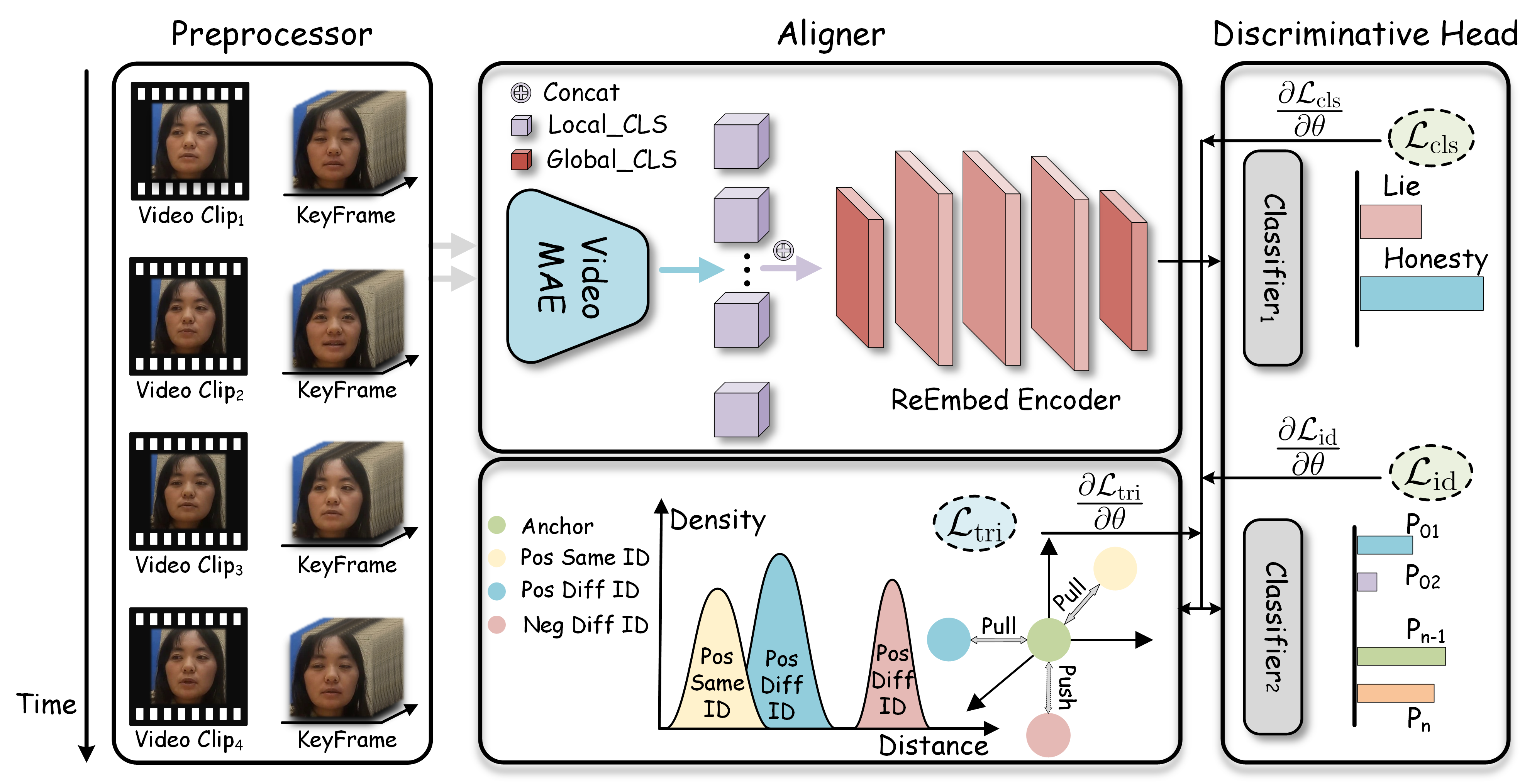}
  \caption{Architecture of the proposed GenLie framework.}
  \label{fig:genlie}
  \vspace{-2ex}
\end{figure*}

To address this core challenge, we propose GenLie, a novel end-to-end framework for video-based lie detection. GenLie adopts a local–global joint modeling strategy: (1) locally, videos are segmented into short clips, where redundancy-aware frame selection suppresses noise and VideoMAEv2 features are refined through a task-driven re-embedding module to highlight latent deceptive cues; (2) globally, adversarial speaker-decorrelation mitigates identity-related interference, while a video-level triplet loss enhances the discriminative structure of the embedding space. The main contributions of this work are summarized as follows:  
\begin{itemize}
  \item We propose GenLie, an end-to-end framework that tackles the fundamental problem of learning sparse yet discriminative representations for video-based detection.  
  \item We introduce a local–global strategy that combines redundancy-aware selection, task-driven re-embedding, and global optimization (adversarial speaker-decorrelation and triplet loss) to produce identity-invariant and discriminative video-level embeddings.  

  \item Extensive experiments on three benchmark datasets demonstrate that GenLie achieves state-of-the-art performance across both high-stakes and low-stakes scenarios.  
\end{itemize}

\section{Methodology}
\subsection{Task Definition}
Let a video sample be denoted as $\mathbf{V} = \{\mathbf{f_t}\}_{t=1}^T$, where $\mathbf{f_t}$ represents the $t$-th frame and $T$ is the total number of frames. The goal of video-based lie detection is to predict a binary label $y \in \{0, 1\}$ indicating whether the speaker is being deceptive ($y = 1$) or truthful ($y = 0$). Formally, the objective is to learn a mapping function:
\vspace{-3ex}
\begin{equation}
    \mathcal{F}: \mathbf{V} \rightarrow \{0, 1\},
\end{equation}
where $\mathcal{F}(\mathbf{V}) = 1$ denotes a deceptive video and $\mathcal{F}(\mathbf{V}) = 0$ denotes a truthful one.

\subsection{Model Overview}

As shown in Figure~\ref{fig:genlie}, GenLie adopts a modular architecture with three components: \textit{Preprocessor}, which segments videos into clips and removes redundant frames; \textit{Aligner}, which extracts latent cues and performs semantic alignment to unify sparse signals into a global representation; and \textit{Discriminative Head}, which conducts video-level classification with auxiliary objectives (speaker-decorrelation and triplet loss) to enforce identity-agnostic and discriminative embeddings.
\vspace{-2ex}
\subsection{Preprocessor}
We divide $\mathbf{V}$ into $N$ equal-length segments ${\mathbf{S_i}}_{i=1}^N$, and from each segment select $K$ informative frames such that $N \times K = 128$ (all frames are preserved if $T < 128$). For each frame $\mathbf{f_t^i}$ in segment $\mathbf{S_i}$, we compute an importance score $\phi(\mathbf{f_t^i})$ and select the top-$K$ frames to form $\mathbf{X_i}$. We explore six different scoring strategies for $\phi(\mathbf{f_t^i})$, three of which—AU-based sampling, micro-expression-based sampling, and gaze-based sampling—are designed based on facial cues proposed in \cite{nam2023facialcuenet}.
\begin{itemize}
    \item \textbf{AU-based:} Prioritizes frames with higher total action unit (AU) intensities from OpenFace.  
    \item \textbf{Micro-expression-based:} Selects frames with more transient AUs ($<$0.5s) to capture brief micro-expressions.  
    \item \textbf{Gaze-based:} Ranks frames by statistical variation of 36D gaze descriptors extracted via OpenFace.  
    \item \textbf{Posture-based:} Chooses frames with larger body keypoint displacements measured by MediaPipe.  
    \item \textbf{Multi-cue Fusion:} Combines normalized AU, micro-expression, gaze, and posture scores for a composite measure.  
    \item \textbf{Uniform:} Evenly samples frames at fixed intervals as a baseline.  
\end{itemize}
\vspace{-3ex}
\subsection{Aligner}
The Aligner receives \( \{\mathbf{X_i}\}_{i=1}^{N} \) from the \textit{Preprocessor}, where each $\mathbf{X_i}$ is the top-$K$ frames of the $i$-th segment. Each segment set is encoded by a frozen VideoMAEv2 encoder $\mathcal{E}(\cdot)$ to obtain its segment-level feature:
\begin{equation}
    \mathbf{h}_i = \mathcal{E}(\mathbf{X_i}), \quad \mathbf{h}_i \in \mathbb{R}^{768}.
\end{equation}

The sequence $\mathbf{H} = [\mathbf{h}_1; \dots; \mathbf{h}_N] \in \mathbb{R}^{N \times 768}$ models short-term dynamics. We then apply mean pooling and a two-layer MLP to derive the final video embedding:
\begin{equation}
    \mathbf{z} = \mathbf{W}_2 \cdot \sigma(\mathbf{W}_1 \cdot \text{MeanPool}(\mathbf{H}) + \mathbf{b}_1) + \mathbf{b}_2,
\end{equation}
where $\sigma(\cdot)$ is ReLU and $\mathbf{z} \in \mathbb{R}^{768}$ serves as the global video representation for downstream tasks.

\vspace{-2ex}
\subsection{Discriminative Head}
Although the semantic aligner provides a compact representation, identity- and context-related noise remains. 
We therefore introduce two complementary strategies.

\textbf{Speaker-Decorrelation.}
To suppress identity-related noise and promote generalization across individuals, we introduce a speaker-decorrelation mechanism based on adversarial training. Specifically, we attach an auxiliary speaker classifier $f_{\text{id}}$ to the global video embedding $\mathbf{z}$ via a Gradient Reversal Layer (GRL). The speaker prediction is computed as:
\begin{equation}
    \mathbf{\hat{y}_{\text{id}}} = f_{\text{id}}(\text{GRL}(\mathbf{z})),
\end{equation}
\begin{equation}
    \mathcal{L}_{\text{id}} = - \sum_{c=1}^{C} \mathds{1}_{[y_{\text{id}}=c]} \cdot \log \mathbf{\hat{y}_{\text{id}}^{(c)}},
\end{equation}
where \( C \) is the number of speaker identities, \( \hat{y}_{\text{id}}^{(c)} \) is the predicted probability for class \( c \), and \( \mathds{1}_{[y_{\text{id}}=c]} \) is the indicator function. 
The GRL forwards $\mathbf{z}$ unchanged but reverses gradients during backpropagation, with a scaling factor \( \lambda \) controlling adversarial strength.

\textbf{Triplet Loss.}
To enhance discriminability, we adopt a video-level \textit{triplet loss} that promotes intra-class compactness and inter-class separation. Given an anchor $\mathbf{z}_a$, a positive $\mathbf{z}_p$ (same label) and a negative $\mathbf{z}_n$ (different label), the loss is:
\begin{equation}
    \mathcal{L}_{\text{tri}} = \max\left(0,\ \|\mathbf{z}_a - \mathbf{z}_p\|_2^2 - \|\mathbf{z}_a - \mathbf{z}_n\|_2^2 + m \right),
\end{equation}
where $m$ is a margin. Positives are sampled both from the same and different speakers, improving robustness to identity variations.

\subsection{Final Optimization Objective}
The overall training loss for the proposed GenLie framework integrates all three components:
\begin{equation}
    \mathcal{L}_{\text{total}} = \mathcal{L}_{\text{cls}} + \alpha \cdot \mathcal{L}_{\text{id}} + \beta \cdot \mathcal{L}_{\text{tri}},
\end{equation}
where $\alpha$ and $\beta$ are weighting factors for the auxiliary objectives. This joint loss guides the model to learn deception-aware, identity-invariant, and discriminative video representations.
\vspace{-2ex}
\section{Experiments}
\vspace{-2ex}
\subsection{Datasets}
We evaluate GenLie on three widely-used deception detection benchmarks, covering both low- and high-stakes scenarios:
\vspace{-2ex}
\begin{itemize}
  \item \textbf{MDPE} \cite{cai2025mdpemultimodaldeceptiondataset}: a large-scale Chinese corpus collected in non-high-stakes conditions, containing 104 hours of video from 193 subjects (1,737 deceptive vs. 2,895 truthful clips). We follow the official train-test protocol.\vspace{-2ex}
  \item \textbf{Real-Life Trial} \cite{perez2015deception}: 121 courtroom videos from trials involving 56 subjects (60 truthful vs. 61 deceptive). We adopt official 121-fold leave-one-out cross-validation.
  \vspace{-2ex}
  \item \textbf{SEUMLD} \cite{xu2024multimodal}: a high-stakes dataset with 3,224 conversational segments from 76 participants (2,112 truthful vs. 1,112 deceptive). We use the original 5-fold protocol.
  \vspace{-2ex}
\end{itemize}
  \vspace{-2ex}
  
We report Positive-class F1-score (F1), Accuracy (ACC), and Area Under the ROC Curve (AUC) as evaluation metrics, all expressed in percentages.
\vspace{-3ex}
\subsection{Implementation Details}
We implement our framework in PyTorch and train on a single NVIDIA RTX 4090 GPU. Adam is used with initial learning rate $1 \times 10^{-5}$ and weight decay $1 \times 10^{-4}$. Batch size is 8. Segment-level features from VideoMAEv2 have dimension 768. The re-embedding MLP has hidden size 1024 and dropout 0.3. For multi-task optimization, weights of 0.1 are applied to both adversarial identity loss and triplet loss, with GRL strength $\lambda=1.0$. All runs are repeated with random seed 42.
\vspace{-2ex}
\subsection{Baselines}
We compare GenLie against representative baselines across three modeling paradigms: \textbf{Explicit-cue methods:} MBF~\cite{rill2019high}, MAD-Net~\cite{bahaa2024advancing}, FTM~\cite{de2024ftm}, and FADe-Net~\cite{ahmed2025deception} rely on handcrafted features such as facial action units, expression dynamics, or micro-expressions to perform deception classification. \textbf{Hybrid methods:} FFCSN~\cite{ding2019face} and FacialCueNet~\cite{nam2023facialcuenet} integrate both handcrafted facial cues and deep spatiotemporal representations through dual-stream or attention-based fusion. \textbf{Implicit-cue methods:} MDPE~\cite{cai2025mdpemultimodaldeceptiondataset}, LM~\cite{zhuo2025lm}, and AFFAKT~\cite{ji2025affakt} adopt end-to-end learning frameworks based on Transformers or vision encoders to extract latent features directly from raw video without relying on predefined cues.

\begin{table*}[t]
\centering
\fontsize{9pt}{9pt} \selectfont
\renewcommand{\arraystretch}{1.2}
\setlength{\tabcolsep}{3.8pt}
\begin{tabular}{lccc ccc ccc}
\toprule[1.2pt]
\multirow{2}{*}{\textbf{Method}} & \multicolumn{3}{c}{\textbf{MDPE}} & \multicolumn{3}{c}{\textbf{Real-Life Trial}} & \multicolumn{3}{c}{\textbf{SEUMLD}} \\
\cmidrule(lr){2-4} \cmidrule(lr){5-7} \cmidrule(lr){8-10}
 & F1 & ACC & AUC & F1 & ACC & AUC & F1 & ACC & AUC \\
\midrule
MBF~\cite{rill2019high} & 0.00 & 60.00 & 48.46 & 0.00 & 49.59 & 0.50 & 13.08 & 64.14 & 51.76 \\
MAD-Net~\cite{bahaa2024advancing} & 1.02 & 60.00 & 51.70 & 91.80 & \uline{91.74} & \uline{98.80} & 2.10 & \textbf{65.32} & 52.44 \\
FTM~\cite{de2024ftm} & 0.00 & 60.00 & 52.93 & 61.67 & 61.98 & 65.79 & \uline{42.08} & 46.65 & 50.05 \\
FADe-Net~\cite{ahmed2025deception} & 0.00 & 60.00 & 48.46 & 0.00 & 49.59 & 0.50 & 13.08 & 64.14 & 51.76 \\
\midrule
FFCSN~\cite{ding2019face} & 37.65 & 57.10 & 53.02 & 90.48 & 90.08 & 97.51 & 41.53 & 62.03 & 56.30 \\
FacialCueNet~\cite{nam2023facialcuenet} & 18.22 & 60.93 & 52.59 & \textbf{93.55} & \textbf{93.39} & 96.56 & 19.53 & 61.41 & 51.92 \\
\midrule
MDPE~\cite{cai2025mdpemultimodaldeceptiondataset} & 10.72 & 60.30 & 57.70 & 88.52 & 88.43 & 96.89 & 37.62 & 63.65 & 55.70 \\
LM~\cite{zhuo2025lm} & 15.08 & 60.31 & 54.06 & 86.18 & 85.95 & 96.69 & 39.64 & 63.06 & \uline{57.04} \\
AFFAKT~\cite{ji2025affakt} & \uline{38.90} & \uline{60.72} & \uline{58.72} & 93.22 & \textbf{93.39} & 72.44 & 0.53 & \uline{65.26} & 50.89 \\
\textbf{Ours (GenLie)} & \textbf{38.97} & \textbf{63.01} & \textbf{59.91} & \uline{93.44} & \textbf{93.39} & \textbf{98.96} & \textbf{42.39} & 64.17 & \textbf{59.15} \\
\bottomrule[1.2pt]
\end{tabular}
\caption{Performance comparison of baselines and GenLie on three datasets.}
\label{tab:main-results}\vspace{-2ex}
\end{table*}

\subsection{Overall Results}
As shown in Table~\ref{tab:main-results}, our proposed method consistently delivers strong performance across all three benchmark datasets. 

Across three benchmarks, GenLie consistently outperforms explicit-cue, hybrid, and implicit baselines. 
On MDPE, it achieves higher F1 and AUC despite class imbalance that makes accuracy less informative. 
On Real-Life Trial, it delivers the best ACC and AUC without relying on spurious low-quality artifacts. 
On SEUMLD, it again tops F1 and AUC under high-stress, adversarial conditions, showing strong robustness and generalization.

\vspace{-2ex}
\subsection{Ablation Study}
As shown in Table~\ref{tab:ablation}, the full GenLie model consistently achieves the best performance. 
\begin{table}[ht]
    \centering
    \fontsize{9pt}{9pt} \selectfont
    \renewcommand{\arraystretch}{1.2}
    \setlength{\tabcolsep}{3pt} 
    \begin{tabular}{l p{0.75cm} p{0.75cm} p{0.75cm} p{0.75cm} p{0.75cm} p{0.75cm}}
        \toprule[1.2pt]
        & \multicolumn{3}{c}{MDPE} 
        & \multicolumn{3}{c}{SEUMLD} \\
        \cmidrule(lr){2-4} 
        \cmidrule(lr){5-7}
        Model & F1 & ACC & AUC 
        & F1 & ACC & AUC \\
        \midrule
        GenLie & \textbf{38.97} & \textbf{63.01} & \textbf{59.91} 
        & \textbf{42.39} & \textbf{64.17} & \textbf{59.15} \\
        w/o TS & 24.95 & 60.10 & 53.49 
        & 38.59 & 57.95 & 51.92 \\
        w/o $\mathcal{F}_{\text{SR}}$ & 18.10 & 60.62 & 58.10 
        & 35.84 & 60.36 & 55.03 \\
        w/o $\mathcal{L}_{\text{id}}$ & 34.06 & 62.28 & 59.78 
        & 36.79 & 57.06 & 53.55 \\
        w/o $\mathcal{L}_{\text{tri}}$ & 25.73 & 62.90 & 58.07 
        & 40.85 & 61.29 & 56.83 \\
        \bottomrule[1.2pt]
    \end{tabular}
\caption{Ablation study results on MDPE and SEUMLD.}
\label{tab:ablation}\vspace{-2ex}
\end{table}
Ablation results show that semantic re-embedding ($\mathcal{F}_{\text{SR}}$) is most crucial, especially on MDPE and SEUMLD, while adversarial debiasing ($\mathcal{L}_{\text{id}}$) is key on Real-Life Trial due to speaker bias. Temporal segmentation (TS) preserves sparse cues, and triplet loss ($\mathcal{L}_{\text{tri}}$) further strengthens discriminability.

\vspace{-2ex}
\subsection{Frame Selection Strategies}
\vspace{-2ex}

\begin{table}[h]
    \centering
    \fontsize{9pt}{9pt} \selectfont
    \renewcommand{\arraystretch}{1.2}
    \setlength{\tabcolsep}{3pt} 
    \begin{tabular}{l p{0.75cm} p{0.75cm} p{0.75cm} p{0.75cm} p{0.75cm} p{0.75cm}}
        \toprule[1.2pt]
        & \multicolumn{3}{c}{MDPE} 
        & \multicolumn{3}{c}{SEUMLD} \\
        \cmidrule(lr){2-4} 
        \cmidrule(lr){5-7}
        Strategy & F1 & ACC & AUC 
        & F1 & ACC & AUC \\
        \midrule
        Uniform & \textbf{38.97} & \textbf{63.01} & \textbf{59.91} 
        & 42.39 & \textbf{64.17} & 59.15 \\
        AU-based & 35.05 & 57.62 & 53.06 
        & 38.14 & 60.76 & \textbf{59.16} \\
        Gaze & 26.87 & 60.52 & 55.49 
        & \uline{42.72} & 58.75 & 57.36 \\
        Micro-exp & 20.73 & 59.59 & 54.46 
        & 42.40 & 57.35 & 57.44 \\
        Posture & 33.04 & \uline{60.93} & \uline{59.16} 
        & 42.11 & 58.56 & 57.75 \\
        Fusion & \uline{37.13} & 58.24 & 53.97 
        & \textbf{42.78} & 57.35 & 57.23 \\
        \bottomrule[1.2pt]
    \end{tabular}
    \caption{Comparison of frame selection strategies on MDPE and SEUMLD.}
    \label{tab:keyframe}
    \vspace{-2ex}
\end{table}
We compare six frame selection strategies. As shown in Table~\ref{tab:keyframe}, uniform sampling achieves the best results across datasets, reflecting its robustness and temporal coverage. Posture-based sampling works as the second choice on MDPE and Real-Life Trial, while gaze-based sampling is more effective on SEUMLD, highlighting dataset cue salience.
\vspace{-2ex}
\subsection{Case Study}
\begin{figure}[ht]
    \centering
    \includegraphics[width=0.96\linewidth]{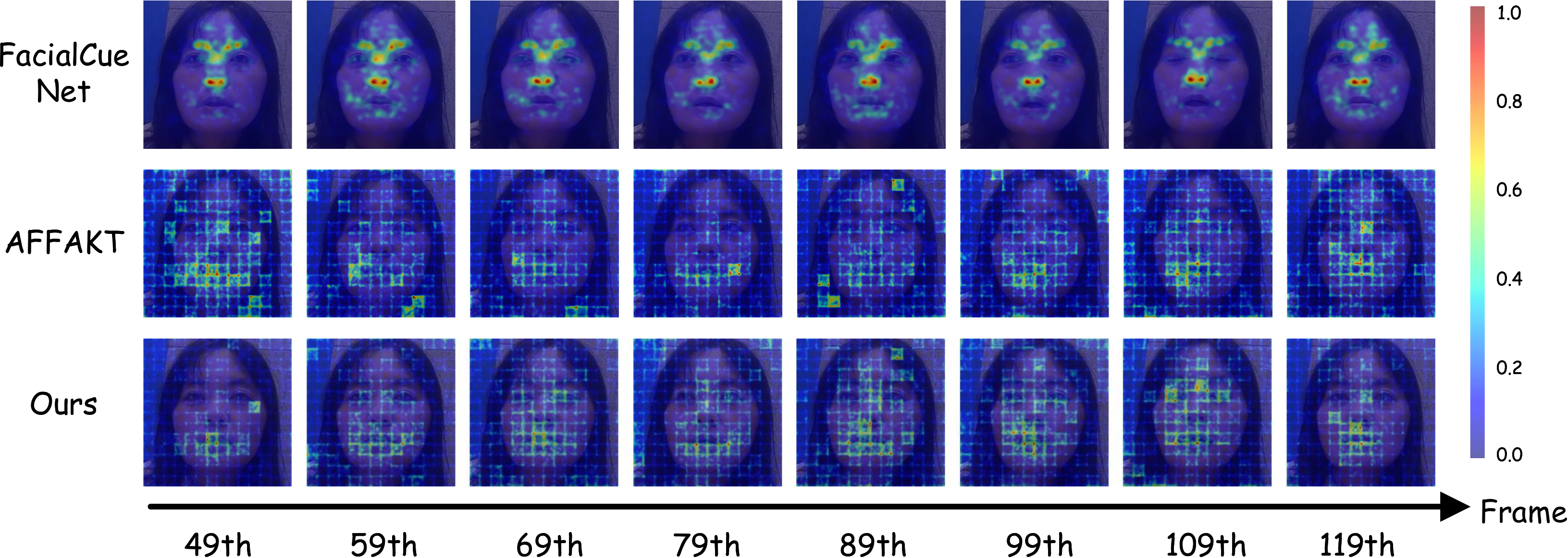}
\caption{Visualization of attention maps from FacialCueNet, AFFAKT, and GenLie on a deceptive clip.}
    \label{fig:case-study}
    \vspace{-2ex}
\end{figure}

\vspace{-1.5ex}
Fig.~\ref{fig:case-study} visualizes attention maps on MDPE. FacialCueNet focuses narrowly on predefined facial regions; AFFAKT scatters attention to irrelevant areas. By contrast, GenLie highlights subtle regions (e.g., mouth corners, nasolabial folds), validating its ability to capture fine-grained cues.
\vspace{-2ex}
\section{Conclusion}
\vspace{-2ex}
We propose GenLie, a framework that performs local feature modeling under global supervision to tackle the challenge of learning sparse, discriminative representations in lie detection. Its core is a local–global strategy: (1) locally, videos are segmented into clips, where redundancy-aware sampling suppresses noise and a task-driven re-embedding module refines VideoMAEv2 features to highlight subtle cues; (2) globally, adversarial speaker-decorrelation reduces identity interference, while a triplet loss strengthens discriminative structure. On MDPE, SEUMLD, and Real-Life Trial, GenLie achieves F1, ACC, and AUC scores surpassing existing methods. Ablations show re-embedding and speaker-decorrelation contribute most, while uniform sampling is more robust than heuristic selection. Overall, GenLie provides a simple, reproducible solution that generalizes across contexts and identities, and integrates with frozen video encoders with minimal tuning.




\bibliographystyle{IEEEbib}
\bibliography{strings,refs}

\end{document}